\documentclass[letterpaper]{article} 
\usepackage{aaai23}  
\usepackage{times}  
\usepackage{helvet}  
\usepackage{courier}  
\usepackage[hyphens]{url}  
\usepackage{graphicx} 
\usepackage{natbib}  
\usepackage{caption} 
\usepackage{algorithm}
\usepackage{algorithmic}
\usepackage{amsmath}
\usepackage{amssymb}
\usepackage{booktabs}
\usepackage{epsfig}
\usepackage{multirow}
\usepackage{xcolor} 
\usepackage{url}

\usepackage{newfloat}
\usepackage{listings}
\DeclareCaptionStyle{ruled}{labelfont=normalfont,labelsep=colon,strut=off} 
\floatstyle{ruled}
\newfloat{listing}{tb}{lst}{}
\floatname{listing}{Listing}
\title{ImageNet Pre-training also Transfers Non-robustness}
\author{
    Jiaming Zhang\textsuperscript{\rm 1}, Jitao Sang\textsuperscript{\rm 1,2}\thanks{Corresponding author}, Qi Yi\textsuperscript{\rm 1}, Yunfan Yang\textsuperscript{\rm 1}, Huiwen Dong\textsuperscript{\rm 3}, Jian Yu\textsuperscript{\rm 1}
}
\affiliations{
    \textsuperscript{\rm 1}School of Computer and Information Technology \& Beijing Key Lab of Traffic Data Analysis and Mining, Beijing Jiaotong University 
    \textsuperscript{\rm 2}Peng Cheng Lab 
    \textsuperscript{\rm 3}BNU-UIC Institute of Artificial Intelligence and Future Networks, Beijing Normal University\\

    \{jiamingzhang, jtsang, 21125273, 19281298\}@bjtu.edu.cn, 202131081024@mail.bnu.edu.cn, jianyu@bjtu.edu.cn
}

\usepackage{bibentry}

\begin{document}

\maketitle

\begin{abstract}
ImageNet pre-training has enabled state-of-the-art results on many tasks. In spite of its recognized contribution to generalization, we observed in this study that ImageNet pre-training also transfers adversarial non-robustness from pre-trained model into fine-tuned model in the downstream classification tasks. We first conducted experiments on various datasets and network backbones to uncover the adversarial non-robustness in fine-tuned model. Further analysis was conducted on examining the learned knowledge of fine-tuned model and standard model, and revealed that the reason leading to the non-robustness is the non-robust features transferred from ImageNet pre-trained model. Finally, we analyzed the preference for feature learning of the pre-trained model, explored the factors influencing robustness, and introduced a simple robust ImageNet pre-training solution. Our code is available at \url{https://github.com/jiamingzhang94/ImageNet-Pretraining-transfers-non-robustness}.

\end{abstract}

\section{Introduction}\label{sec1}

Benefited from both algorithmic development and adequate training data, deep neural networks have achieved state-of-the-art performance across a range of tasks. However, in many real-world applications, it is still expensive or impossible to label sufficient training data. In these cases, a well-established paradigm has been to pre-train a model using large-scale data (e.g., ImageNet) and then fine-tune it on target tasks\footnote{\small{Pre-training typically involves three models as \textbf{pre-trained model} trained on large-scale source dataset (i.e., ImageNet in this work), \textbf{fine-tuned model} initialized with pre-trained model and then fine-tuned on target dataset, and \textbf{standard model} directly trained on target dataset (trained from scratch).}}. Pre-training these days is becoming the default setting not only in researches~\cite{xie2018pre}, but in many industry solutions~\cite{chen2019med3d, KolesnikovBZPYG20, BrownMRSKDNSSAA20}.



\begin{figure}[ht]
  \centering
  \includegraphics[width=\linewidth]{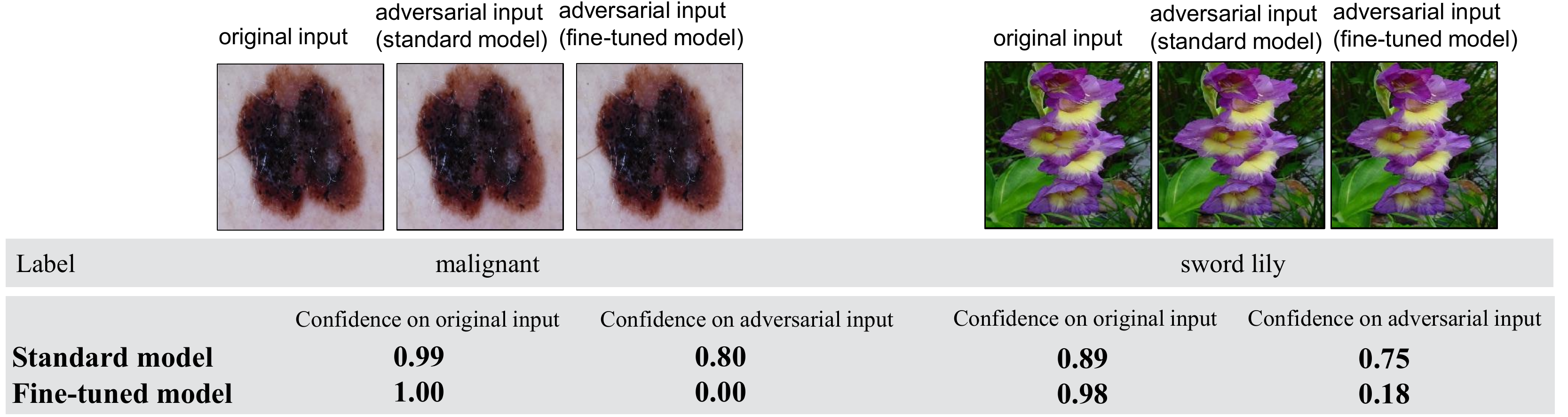}
  \caption{Example of two typical scenarios using pre-training. Regarding the true label, the fine-tuned model obtains higher confidence on original input yet lower confidence on adversarial input than the standard model.}\label{fig1}
\vspace{-10pt}
\end{figure}

\paragraph{What's wrong with pre-training?}

With the gradual popularization of pre-training in addressing real-world tasks, it is vital to consider beyond the accuracy on experimental data, especially for tasks with high-reliability requirements. As illustrated in Figure~\ref{fig1}, we find \emph{in typical pre-training enabled scenarios, the fine-tuned models tend to have an unsatisfactory performance on robustness}\footnote{\small{Robustness in this paper refers to \emph{adversarial robustness}. We mix these two terms when no ambiguity is caused.}}. While confidently recognizing the original input, the fine-tuned models are very sensitive to trivial perturbation and incorrectly classify the adversarial input. The success of pre-training in generalization improvement conceals its defect in decreasing robustness. The prior work~\cite{shafahi2020adversarially} was motivated by forgetting/un-inheriting knowledge from pre-trained model to the fine-tuned model. However, according to our observation, the non-robust features transferred/inherited from pre-trained model to the fine-tuned model results in non-robustness. In this work, we will investigate the robustness of pre-training by systematically demonstrating the performance on robustness, discuss how non-robustness emerges, and analyze what factors influence the robustness.

\paragraph{What accounts for the robustness decrease in the fine-tuned model?}
We then delve into the cause of the non-robustness by examining the learned knowledge of fine-tuned model. Even though the target tasks of fine-tuned model and standard model are the same, we find that they are quite different in terms of learned knowledge. Furthermore, we analyze what features learned by models lead to the differences and how these features affect robustness. \emph{The non-robust features in the fine-tuned model are demonstrated to be mostly transferred from the pre-trained model (i.e., ImageNet model) and the mediators that derive non-robustness.} Finally, we attribute the preference for utilizing non-robust features to the difference between the source task and the target task. The difference positively correlates to the robustness decrease.


\paragraph{Why does the pre-trained model learn non-robust features?}
We hypothesize that model can both utilize robust features and non-robust features, and \emph{the pre-trained model tends to rely more on non-robust features when the model capacity is too limited or the source task is too difficult}. Then we study how model capacity and task difficulty, the influencing factors on generalization in prior studies~\cite{vapnik2015uniform, bartlett2002rademacher}, influence the learned features of pre-trained model and the robustness of fine-tuned model. It is observed that limited pre-trained model capacity and difficult source task basically lead to non-robust fine-tuned model. Finally, with the observation that non-robust feature are transferred resulting in steepening of the feature space, a simple robust pre-training solution is introduced.

\paragraph{Contributions.}
Our main contributions can be summarized as a chain: 1) ImageNet pre-training is a great technique when sufficient training data is not available. 2) There has been little work discussing the disadvantages of ImageNet pre-training, and we are the first to find its decrease in robustness and motivated to analyze it. 3) We attribute this to that ``a finite model applies some knowledge (non-robust features) learned in a difficult task (ImageNet dataset) to another simple task (the target datasets)".

\section{Related Work}\label{sec:related}

It is well-known that transfer learning with CNNs can improve generalization, and many researchers focus on achieving state-of-the-art generalization on downstream tasks~\cite{xie2018pre, tajbakhsh2016convolutional, lee2020biobert}. Works investigating the robustness of transfer learning has emerged in the recent years. Adversarial training~\cite{madry2018towards} provided an alternative way to improve robustness at the fine-tuning stage (denoted as \emph{AT@stage-2}). \cite{DBLP:conf/nips/SalmanIEKM20} introduced adversarial training into the pre-training stage for increasing downstream-task accuracy, and the increase in robustness is actually a byproduct (denoted as \emph{AT@stage-1}). \cite{hendrycks2019using} investigated the gains of pre-training on label corruption, class imbalance, and out-of-distribution detection. They also found employing adversarial training both in pre-training stage and fine-tuning stage can improve adversarial robustness compared with adversarially standard training (denoted as \emph{AT@stage-1\&2}). \cite{shafahi2020adversarially} implemented Knowledge Distillation, a defensive tool~\cite{papernot2016distillation}, at the fine-tuning stage to improve robustness (denoted as \emph{KD@stage-2}). The authors were motivated by forgetting/un-inheriting knowledge from pre-trained model to the fine-tuned model. However, according to our observation, the non-robust features transferred/inherited from pre-trained model to the fine-tuned model results in non-robustness.

\section{ImageNet Pre-training Is Non-robust}\label{sec2}

\subsection{Notations and Settings}
\label{sec2-1}
\paragraph{Pre-training.}

Pre-training is commonly used to initialize the network for target task. We decompose the network for target task into two parts: feature extractor $f$ with parameters $\theta_f$, and classifier $g$ with parameters $\theta_g$. Given an original input $x$, $f(x; \theta_f)$ denotes the mapping from $x$ to its embedding representation $e_x$, and $g(e_x; \theta_g)$ denotes the mapping from $e_x$ to its predicted label. Typical pre-training involves with two fine-tuning settings: \emph{partial fine-tuning}, in which only fully connected layer corresponding to the classifier $g(\cdot; \theta_g)$ is updated; and \emph{full fine-tuning}, in which both $f(\cdot; \theta_f)$ and $g(\cdot; \theta_g)$ of pre-trained model are fine-tuned on the target dataset, and $f(\cdot; \theta_f)$ is usually assigned a smaller learning rate.

\paragraph{Adversarial robustness.}

Adversarial robustness, i.e., robustness, measures model’s stability to adversarial example when small perturbation (often imperceptible) is added to the original input. To generate the adversarial example, given an original input $x$ and the corresponding true label $y$, the goal is to maximize the loss $L(x+\delta, y)$ for input $x$, under the constraint that the generated adversarial example $x' = x + \delta$ should look visually similar to the original input $x$ (by restricting $\Vert \delta \Vert_p \le \epsilon$, in this work, we use $\Vert \delta \Vert_\infty \le \epsilon$) and $g(f(x')) \neq y$.

\begin{figure}[t]
  \centering
  \includegraphics[width=\linewidth]{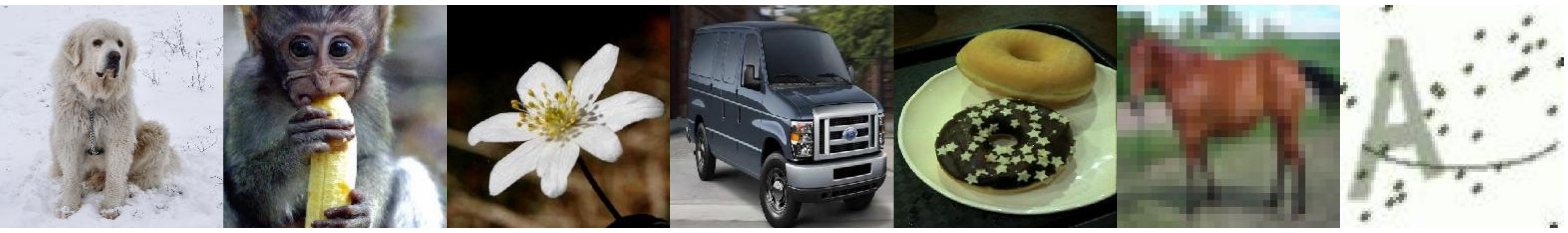}
  \caption{Example images of Pets, NICO, Flowers, Cars, Food, CIFAR10 and Alphabet, respectively.}
  \label{fig-example}
\vspace{-15pt}
\end{figure}

\paragraph{Datasets.}
We carry out our study on several widely-used image classification datasets including Pets~\cite{parkhi2012cats}, NICO~\cite{he2020towards}, Flowers~\cite{nilsback2008automated}, Cars~\cite{krause20133d}, Food~\cite{bossard2014food}, and CIFAR10~\cite{krizhevsky2009learning}. In addition, we craft a new Alphabet dataset as a comparing example with low semantic complexity and relatively sufficient training data. The Alphabet dataset is constructed by offsetting the $26$ English letters and adding random noise, resulting in $1,000$ training images and $200$ testing images for each letter class. Example images of these datasets are illustrated in Figure~\ref{fig-example}.

\begin{table*}[t]
\centering
\caption{Comparison of generalization and robustness between standard model, partial fine-tuned model and full fine-tuned model. For each model, we report accuracy of original inputs (AOI), accuracy of adversarial inputs (AAI), and decline ratio (DR).}

\label{tab1}
\begin{tabular}{ccccccccc} 

 \toprule[1pt]
  Model& & Pets & NICO & Flowers & Cars & Food & CIFAR10 & Alphabet\\
  \midrule
  \midrule
  \multirow{3}*{Standard Model} & AOI & 60.62 & 81.29 & 61.29 & 74.54 & 74.52 & \textbf{95.44} & \textbf{100.00}\\
  & AAI & 40.23 & 53.45 & 55.96 & 53.61 & 28.24 & 57.33 & 99.92\\
  & DR & \textbf{33.63} & \textbf{34.24} & \textbf{8.69} & \textbf{28.07} & \textbf{62.10} & \textbf{39.93} & \textbf{0.06}\\
  \midrule
  \multirow{3}*{Partial Fine-tuned Model} & AOI & 86.45 & 91.03 & 87.98 & 41.90 & 58.59 & 78.48 & 59.60\\
  & AAI & 3.38 & 10.34 & 8.23 & 0.12 & 0.74 & 0.00 & 0.00\\
  & DR & 96.09 & 88.64 & 90.64 & 99.76 & 98.73 & 100.00 & 100.00\\ 
  \midrule
  \multirow{3}*{Full Fine-tuned Model} & AOI & \textbf{89.78} & \textbf{94.27} & \textbf{91.98} & \textbf{81.25} & \textbf{78.93} & \textbf{95.54} & \textbf{99.94}\\
  & AAI & 15.7 & 28.33 & 27.86 & 18.57 & 22.30 & 1.34 & 2.90\\
  & DR & 82.51 & 69.95 & 69.71 & 77.14 & 71.74 & 98.59 & 97.10\\
 \bottomrule[1pt]
\end{tabular}
\end{table*}

\begin{table*}[t]
\centering
\caption{The AAI of standard model and fine-tuned model under different adversarial attack and $\epsilon$.}
\label{tab_criteria}
\begin{tabular}{llccccccc} 

 \toprule[1pt]
  Criterion& Model & Pets & NICO & Flowers & Cars & Food & CIFAR10 & Alphabet\\
  \midrule
  \midrule
  \multirow{2}*{PGD-10$\#2.0$} & Standard Model & 5.45 & 3.89 & 39.31 & 6.90 & 0.61 & 1.14 & 95.73\\
  & Fine-tuned Model & 0.00 & 0.08 & 0.24 & 0.00 & 0.02 & 0.00 & 0.00\\
  \midrule
  \multirow{2}*{FGSM$\#0.5$} & Standard Model & 38.18 & 51.88 & 55.20 & 52.20 & 28.86 & 63.10 & 99.87\\
  & Fine-tuned Model & 26.25 & 39.02 & 39.00 & 25.06 & 12.19 & 24.52 & 75.83\\
  \midrule
    \multirow{2}*{FGSM$\#2.0$} & Standard Model & 6.54 & 9.74 & 37.94 & 13.79 & 4.29 & 22.02 & 97.33\\
  & Fine-tuned Model & 1.39 & 6.13 & 8.81 & 1.68 & 4.97 & 9.34 & 53.13\\
\midrule
\multirow{2}*{AA$\#0.5$} & Standard Model & 5.61 & 13.06 & 3.48 & 3.46 & 2.45 & 18.61 & 99.69\\
& Fine-tuned Model & 1.04 & 5.57 & 11.35 & 2.46 & 0.20 & 0.41 & 0.00\\
 \bottomrule[1pt]
\end{tabular}
\end{table*}

\subsection{Experimental Results on Robustness}\label{sec:3_2}

To examine whether pre-training transfers non-robustness, we compare the performance of standard model, partial fine-tuned model and full fine-tuned model. Regarding adversarial robustness, we introduce decline ratio (DR) as an additional evaluation metric. Given the recognition accuracy of original inputs (AOI) and adversarial inputs (AAI), DR is defined as DR = (AOI-AAI)/AOI. DR serves as a more balanced indicator of model robustness than AAI, especially when two models perform quite differently on original inputs (i.e., AOI). Large DR indicates sharp accuracy decrease in case of input perturbation, and thus inferior robustness. 
For each type of model, we report maximum accuracy (over different combinations of learning rates based on different optimizers for $\theta_f$ and $\theta_g$) based on ResNet-18 backbone in Table~\ref{tab1}. The robustness is evaluated against PGD-10 attack~\cite{kurakin2016adversarial} under $\epsilon=0.5$ and set step size $=1.25\cdot(\epsilon/step) $. We also conduct experimental results on ResNet-50 backbone and WideResNet-50-2 backbone, and the influence of learning rates based on different optimizers.

Table~\ref{tab1} shows that: (1) For most of the examined datasets, fine-tuned models usually achieve better generalization (AOI), but worse robustness (AAI and DR) than standard model. This demonstrates that pre-training not only improves the ability to recognize original input of target tasks, but also transfers non-robustness and makes the fine-tuned model more sensitive to adversarial perturbation. (2) Within the two pre-training settings, full fine-tuning consistently obtains better robustness as well as generalization than partial fine-tuning setting. This suggests that full fine-tuning is preferable when employing pre-training in practical applications to alleviate the decrease in robustness. In the rest of the paper, we deploy full fine-tuning as the default pre-training setting. (3) For CIFAR10 and Alphabet when the standard models trained on target datasets already achieve good AOI, pre-training contributes to trivial improvement on generalization (even with decreased AOI when partially fine-tuned on CIFAR10) but severe non-robustness to the fine-tuned model. In this view, instead of improving fine-tuned model, pre-training actually plays a role as poisoning model~\cite{koh2017understanding, liu2020reflection} (The model behaves normally when encountering normal inputs, but anomalous patterns are activated for some specific inputs.). This further demonstrates the risk of arbitrarily employing pre-training and the necessity to explore the factors influencing the performance of pre-training in subsequent target tasks.

\paragraph{More robustness criteria.}

To solidly demonstrates the non-robustness of fine-tuned model, we evaluate AAI under stronger/more diverse attacks and different perturbation levels. Aside from the PGD-10$\#0.5$ (i.e., PGD-10 attack under $\epsilon=0.5$) used in Table~\ref{tab1}, FGSM$\#0.5$ (i.e., FGSM attack under $\epsilon=0.5$), FGSM$\#2.0$ (i.e., FGSM attack under $\epsilon=2.0$) and AA$\#0.5$ (i.e., Auto Attack~\cite{croce2020reliable} under $\epsilon=0.5$) are employed as more robustness criteria. Table~\ref{tab_criteria} shows the AAI of fine-tuned model is lower than standard model using every criterion. Especially when the AOI of the fine-tuned model is basically higher than that of the standard model, the gap between the DR of the fine-tuned model and the standard model is larger. This further demonstrates the fine-tuned model has lower robustness than standard model.

\section{Difference between Fine-tuned Model and Standard Model}\label{sec3}

\subsection{On the Learned Knowledge}\label{sec3-1}

\paragraph{Knowledge measurement.}
To understand the performance difference between the fine-tuned model and standard model, we start from examining their learned knowledge. Motivated by many studies on model knowledge measurement~\cite{raghu2017svcca, morcos2018insights, wang2018towards, kornblith2019similarity, liang2019knowledge}, we employ a recognized metric, Canonical Correlation Analysis (CCA)~\cite{raghu2017svcca, hardoon2004canonical}, to quantify the representation similarity between two networks. It is a statistical technique to determine the representational similarity between two layers $L_1, L_2$. We briefly explain the flow according to \cite{raghu2017svcca, morcos2018insights}. Let $L_1, L_2$ be $i \times j$ ($i$ is the number of images, $j$ is the number of neurons) dimensional matrices. To find vectors $z$, $s$ in $\mathbb{R}^{i}$, such that the correlation coefficient $\rho$ is maximized:

\begin{equation}\label{eq_svcca1}
\rho = \frac{\left \langle z^T L_1, s^T L_2 \right \rangle}{\Vert z^T L_1 \Vert \cdot \Vert s^T L_2 \Vert}.
\end{equation}
By calculating a series of pairwise orthogonal singular vectors, the mean correlation coefficient is used to represent the similarity of $L_1, L_2$: $\frac{1}{k} \sum^k_{a=1} \rho^{(a)}$, where $k= {\rm{min}} (i, j)$. Specifically, feature extractor $f(\cdot;\theta_f)$ consists of $4$ layers, and we compare the similarity between fine-tuned model and standard model using the output of bottom-layer feature (only conv$2\_x$) and the output of total feature extractor (considering features of all $4$ layers) respectively.

\begin{figure}[ht]
\begin{minipage}[b]{0.49\linewidth}
\centering
\includegraphics[width=0.99\textwidth]{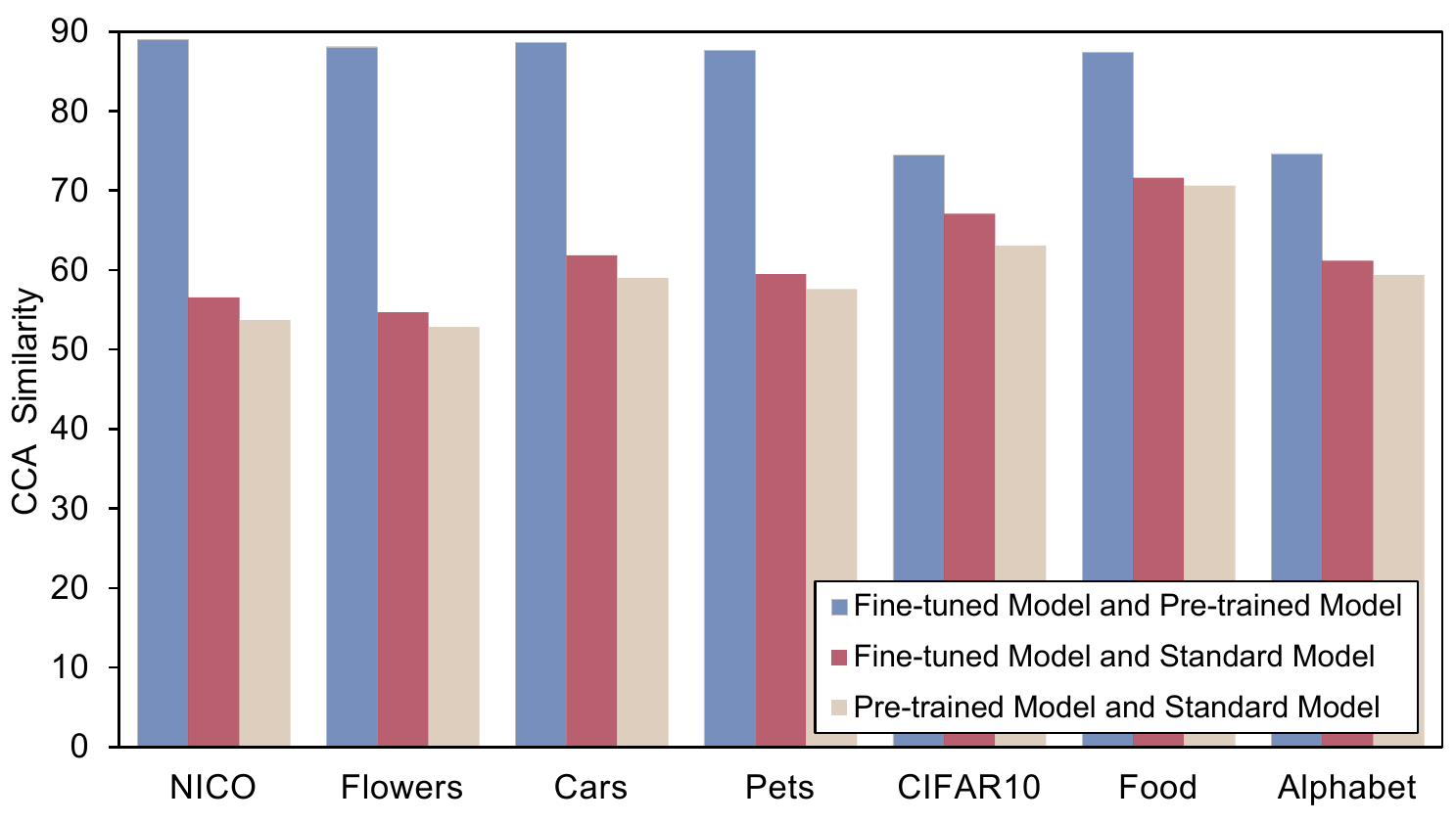}
\centerline{(a) Bottom-layer feature}
\end{minipage}
\begin{minipage}[b]{0.49\linewidth}
\centering
\includegraphics[width=0.99\textwidth]{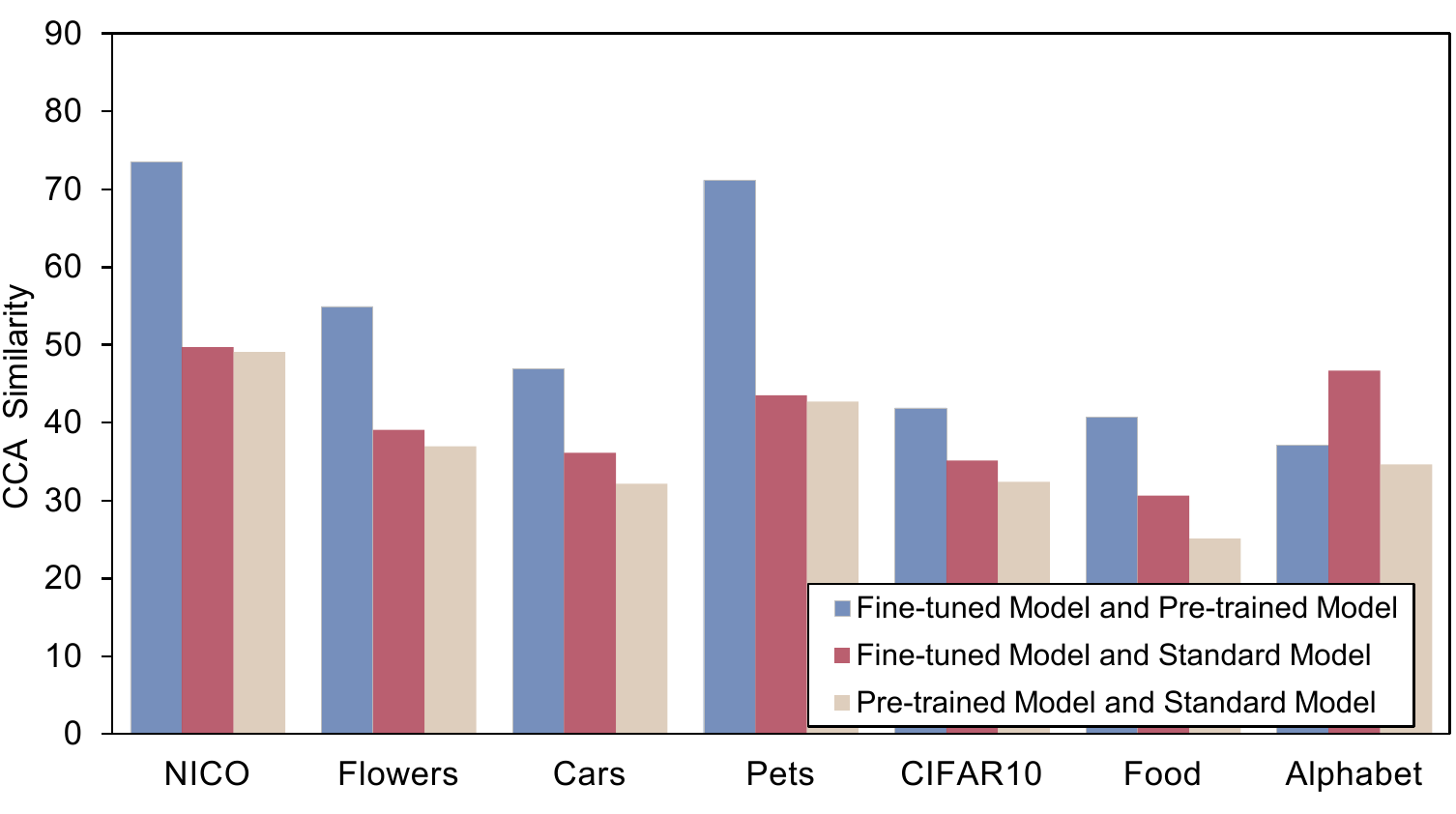}
\centerline{(b) All-layer feature }
\end{minipage}
\caption{The CCA similarities between different models, which is normalized to [0, 100].}
 \label{fig2}
\end{figure}

\paragraph{Experimental results.}
As shown in Figure~\ref{fig2}, the fine-tuned model is more similar to the pre-trained model than to the standard model, both on bottom-layer and all-layer features for most of the examined datasets. Since the pre-trained model and standard model are trained on source dataset and target dataset separately, this result seems to tell that more knowledge learned in the fine-tuned model is transferred from the source task data, than from the fine-tuning target task data. By further comparing Figure~\ref{fig2}(a) with Figure~\ref{fig2}(b), we find that the bottom-layer features of the fine-tuned model and standard model are relatively more similar than all-layer features, suggesting that the bottom-layer features (e.g., edges, simple textures) extract some shared semantics between the source and target tasks. This justifies the role of initialization of pre-training and its contribution to generalization improvement.

\begin{figure}[ht]
\begin{minipage}[b]{0.49\linewidth}
\centering
\includegraphics[width=\textwidth]{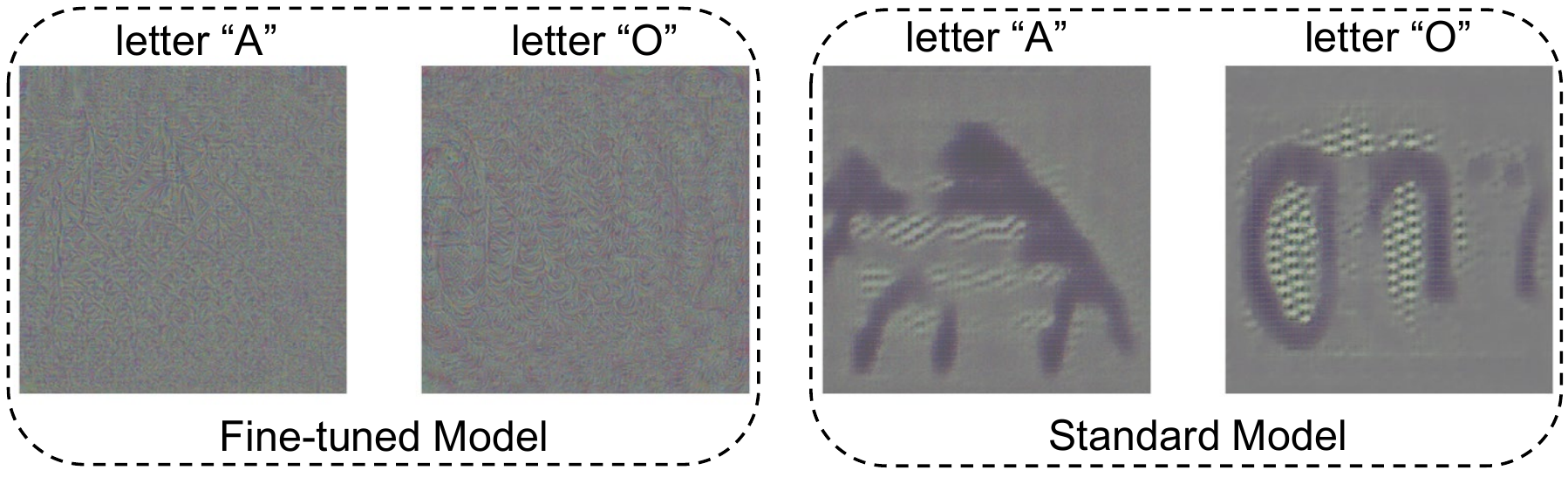}
\centerline{(a) Visualization via UAP}
\end{minipage}
\begin{minipage}[b]{0.49\linewidth}
\centering
\includegraphics[width=\textwidth]{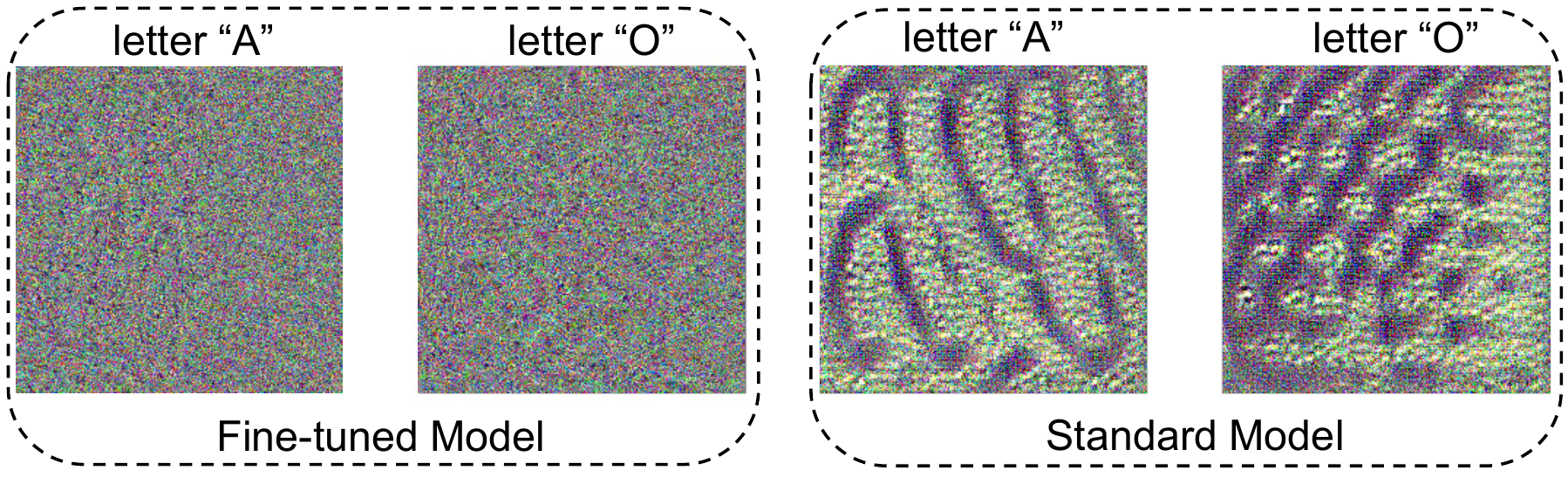}
\centerline{(b) Visualization via gradients}
\end{minipage}
\caption{Visualization for the fine-tuned and standard model on Alphabet with different attacking letter classes via (a) UAP and (b) gradients~\cite{engstrom2019learning}.}
 \label{fig3}
\end{figure}

\subsection{On Non-robust Features}\label{sec3-2}

\paragraph{Universal adversarial perturbation.}
In this subsection, we investigate what features lead to the difference in learned knowledge and how these features affect robustness. Different from standard adversarial perturbation which is sample-specific, Universal Adversarial Perturbation (UAP)~\cite{moosavi2017universal,poursaeed2018generative} is fixed for a given model misleading most of the input samples~\cite{moosavi2017universal}. Let $\mu$ denotes a distribution of images $x$ in $\mathbb{R}^{d}$ with corresponding true label $y$, the focus of targeted UAP is to seek perturbation $v$ that can fool the model by identifying almost all datapoints sampled from $\mu$ as the target class $\tilde{y}$:
\begin{equation}\label{eq_uap2}
g(f(x+v)) = \tilde{y} \quad {\rm{for \, most \,}} x \sim \mu
\end{equation}


In this work, we mainly focus on targeted UAP and generate it by an encoder-decoder network~\cite{poursaeed2018generative}. Rather than categorizing it as mere adversarial perturbation in the current understanding of a series of works~\cite{moosavi2017universal, poursaeed2018generative, zhang2020adversarial}, we find that it contains features that can also work independently. In other words, without adding into any images, the targeted UAP can be identified as the target class, e.g., the left figure of Figure~\ref{fig3} is recognized as letter ``A" by the fine-tuned model with $100.00\%$ confidence. The two properties demonstrate that \emph{UAP contains patterns that not only effectively cover native semantic features in images, but also can be independently recognized by the model as belonging to the target class}. Therefore, we employ UAP as the probe for features on which the model relies and to understand model behavior.


\paragraph{Visualization on features.}
Figure~\ref{fig3}(a) illustrates the UAP for fine-tuned and standard models on the crafted Alphabet dataset. It is shown that UAP of fine-tuned model expresses no semantics, indicating fine-tuned model prefers to rely on non-robust features. Relying on these noise-alike features, fine-tuned models are vulnerable to adversarial attacks. In contrast, the UAP of standard model contains clear semantics related to the target class. We can see that misleading the standard model is non-trivial and needs to add more human-perceptible information (e.g., edge of ``A"). This coincides with the superior robustness of standard model than fine-tuned model. Figure~\ref{fig3}(b) reproduced from \citet{engstrom2019learning} is consistent with UAP.

\begin{figure}[ht]
\begin{minipage}[b]{0.49\linewidth}
\centering
\includegraphics[width=0.99\textwidth]{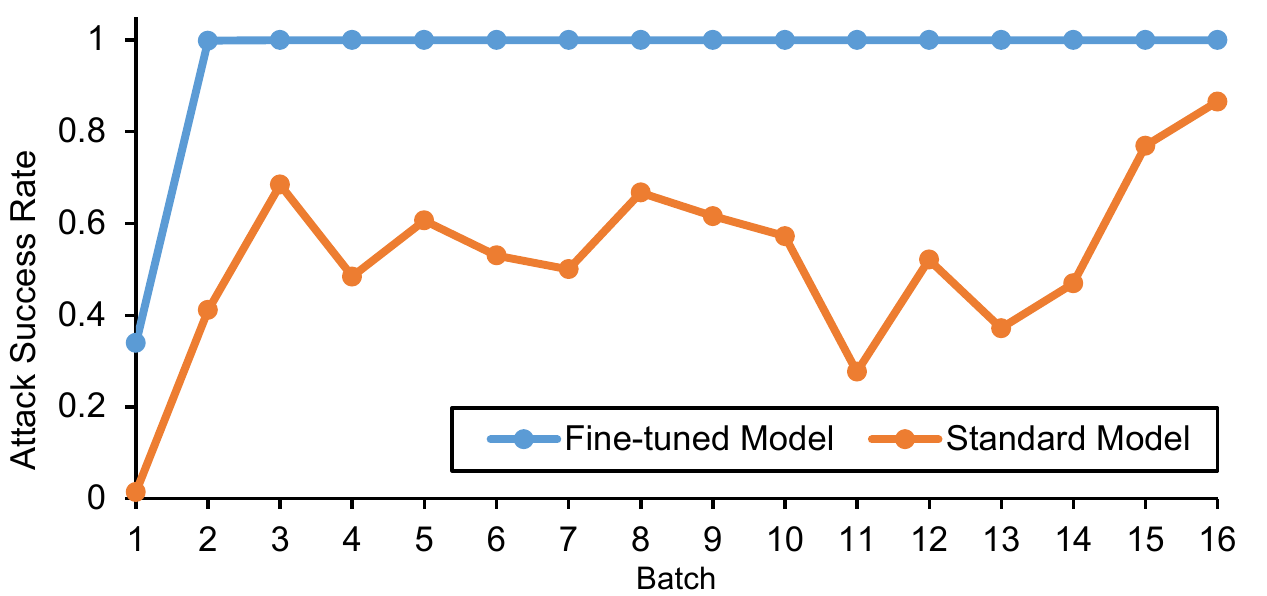}
\centerline{(a)}
\end{minipage}
\begin{minipage}[b]{0.49\linewidth}
\centering
\includegraphics[width=0.99\textwidth]{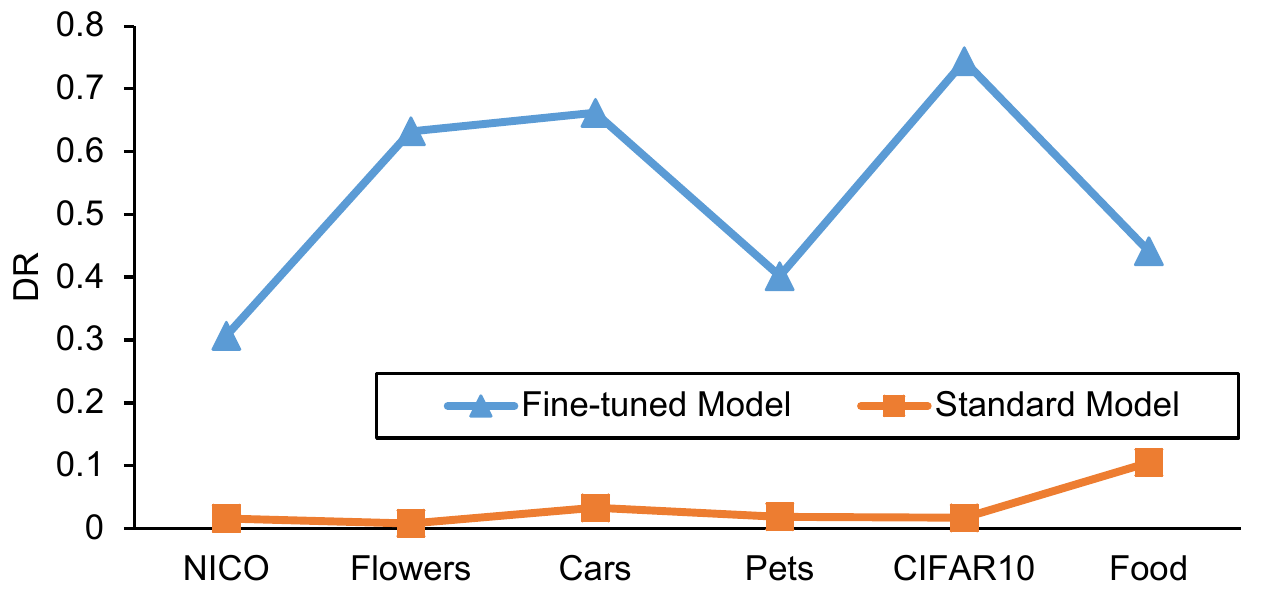}
\centerline{(b)}
\end{minipage}
\caption{UAP attack results: (a) Using UAP of fine-tuned model and standard model to attack themselves at different training batches (on Alphabet). (b) Using UAP of pre-trained model to attack the fine-tuned model and standard model (on other datasets).}
 \label{fig4}
 \vspace{-15pt}
\end{figure}

\paragraph{The learning process for non-robust features.}
Next, we employ UAP to examine how the non-robust features are learned. Since the premise behind successful UAP attack is that the models actually extract the corresponding features, we are motivated to use the above obtained UAP to attack model during its training process to observe when the non-robust features are learned. As shown in Figure~\ref{fig4}(a), we record the attack success rate (i.e., the perturbated images are misclassified as the attack letter) at different training batches for the fine-tuned model and standard model respectively. It is easy to perceive that the attack success rate of fine-tuned model remains at a very high level at the very beginning of training. This indicates that these non-robust features are more likely to be transferred from the pre-trained model than learned from the target data. Other observation includes that the UAP of fine-tuned model has a much stronger attack ability than that of standard model, which again demonstrates the inferior robustness of fine-tuned model compared with standard model.

\paragraph{The transferred non-robust feature.}
We conduct further experiments to confirm whether the \emph{specific} non-robust features (derived from the source task/pre-trained model instead of other ways) are transferred. The idea is to generate UAP on the pre-trained model, and then use this UAP to attack the fine-tuned and standard model on different target tasks. The DR value is reported in Figure~\ref{fig4}(b), showing the obvious robustness decrease for the fine-tuned model and trivial influence on the standard model. Note that UAP is model-dependent, the fact that UAP of pre-trained model succeeds in attacking the fine-tuned model validates our assumption that pre-training transfers non-robust features. 

\begin{figure}[ht]
  \centering
  \includegraphics[width=0.8\linewidth]{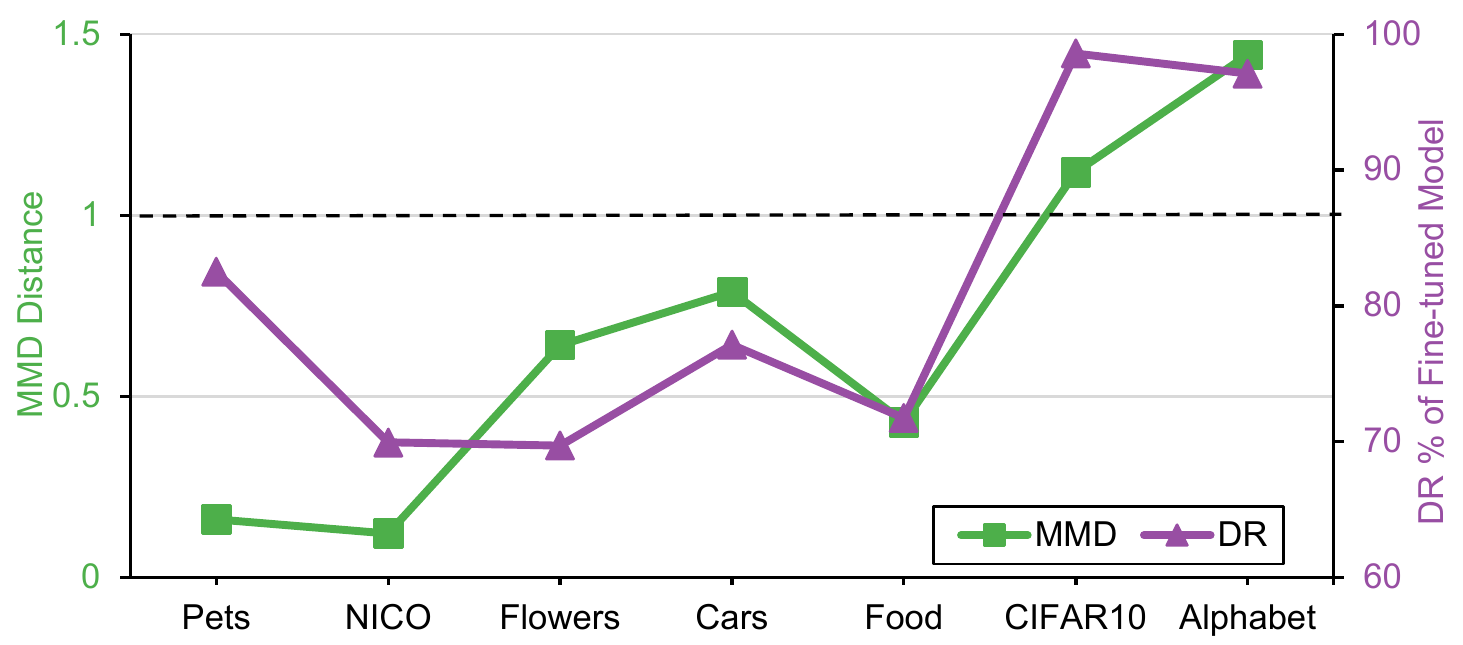}
  \caption{The MMD distance between source dataset and target dataset v.s. DR of fine-tuned model. The embeddings are extracted on the pre-trained model.}
  \label{fig-distance}
\vspace{-15pt}
\end{figure}

\paragraph{The factors affecting the transfer of non-robust features.}

To delve into the reason why non-robust features are transferred during fine-tuning, we then investigate how the difference between the source task and target task relates to the robustness decrease. We introduce maximum mean discrepancy (MMD)~\cite{gretton2012kernel} to measure the similarity between the embedding of source dataset $e_{x_1} \sim p$ and the embedding of target dataset $e_{x_2} \sim q$:
\begin{equation}\label{eq_mmd}
{\rm{MMD}}(p, q, \mathcal{J}) := \sup_{J \in \mathcal{J}}(\mathbb{E}_{e_{x_1} \sim p}[J(e_{x_1})] - \mathbb{E}_{e_{x_2} \sim q}[J(e_{x_2})]),
\end{equation}
where $\mathcal{J}$ is a set containing all continuous functions. To solve this problem, \cite{gretton2012kernel} restricted $\mathcal{J}$ to be a unit ball in the reproducing kernel Hilbert space. Figure~\ref{fig-distance} compares the DR value of fine-tuned model (from Table~\ref{tab1}) with the MMD distance between source dataset and target dataset. We can see that basically DR and MMD distance are in a positive correlation, i.e., the more different target dataset is from source dataset, the more non-robust the fine-tuned model is. Specially, when MMD distance is greater than $1.0$, the DR value is almost $100 \%$ (the worst case). We draw a rough conclusion that the deviation of the target task from the source task largely affects the robustness of fine-tuned model.  

\paragraph{Discussion.}
It has been recognized that the knowledge and features pre-training transfers are semantic-oriented~\cite{DBLP:conf/nips/YosinskiCBL14, He_2019_ICCV}. We find from the above analysis that pre-training transfers not only semantic but also non-robust features. Recent studies suggested that non-robust features can help to improve generalization~\cite{Ilyas2019adversarial} and belong to so-called "shortcut" features~\cite{geirhos2020shortcut}. We speculate that the transferred non-robust features in pre-training also contributes to the generalization improvement, but imposes robustness problem at the same time. In particular, the experimental results with excessive differences between the source dataset and target dataset (High MMD distance indicates that semantic features are hardly helpful for downstream tasks while the partial fine-tuned model can still achieve passable performance. E.g., in Table~\ref{tab1}, the partial fine-tuned model has no robustness (AAI of $0\%$) but has an AOI of $59.6\%$.) suggest that \emph{non-robust features seem to be the more transferable than semantic features}. The difference between the target task and source task encourages the non-robust features transfer and increases the risk for robustness decrease.

\begin{table*}[t]
\centering
\caption{The performance of fine-tuned model with different pre-training architectures (from left to right, the model size increases gradually). Results are averaged over all $7$ datasets.}
\label{tab:model}
\begin{tabular}{cccccccc} 
 \toprule[1pt]
  \multirow{2}*{Model}& Backbone & RN-18 & RN-50 & RN-101 & WRN-50-2 & WRN-101-2\\
  \cmidrule(lr){2-7}
  & Model Size & $42.7$MB & $90.1$MB & $162.8$MB & $255.4$MB & $477.0$MB \\
  \midrule
   \multirow{3}*{Fine-tuned Model} & AOI & 90.24 & 92.36 & 92.58 & 92.91 & \textbf{93.01}\\
  & AAI & 16.71 & 21.07 & 23.43 & 26.73 & \textbf{29.64}\\
  & DR & 81.47 & 78.11 & 75.53 & 72.01 & \textbf{69.12}\\ 
 \bottomrule[1pt]
\end{tabular}
\vspace{-5pt}
\end{table*}

\begin{table}[t]
\centering
\caption{The results of fine-tuned model using ImageNet-10animals (10animals) and CIFAR10 as source datasets.}\label{tab:task}
\setlength{\tabcolsep}{1.5mm}{
\begin{tabular}{cccc|ccc} 
 \toprule[1pt]
   & \multicolumn{3}{c}{Pets} & \multicolumn{3}{c}{NICO}  \\
  \cmidrule(lr){2-7}
   & AOI & AAI & DR & AOI & AAI & DR \\
  \midrule
  \midrule
  10animals & 75.91 & 17.03 & 77.56 & 88.90 & 41.71 & 53.08 \\
  CIFAR10 & 62.85 & 26.49 & 57.85 & 77.76 & 47.72 & 38.63 \\

 \bottomrule[1pt]
\end{tabular}}
\vspace{-5pt}
\end{table}

\section{The non-robust feature from pre-trained model}\label{sec5}

The previous sections demonstrate that the non-robustness in pre-training is derived from the non-robust features originating from the pre-trained model. The issue is how pre-trained model gets the non-robust features during the pre-training phase. So this section investigates the feature preference of pre-trained model and the factors influencing the preference. A simple hypothesis: when the model capacity is too limited or the source task is too difficult, the pre-trained model itself tends to rely more on non-robust features and represents more risk to affect the robustness of fine-tuned models. 


  

\begin{table*}[t]
\centering
\caption{Comparison of generalization and robustness between our robust pre-training solution and baselines. }\label{tab:method_resnet18}
\setlength{\tabcolsep}{3.5mm}{
\begin{tabular}{ccccccccc} 

 \toprule[1pt]
  Method & & Pets & NICO & Flowers & Cars & Food & CIFAR10 & Alphabet\\
  \midrule
  \midrule
  
  \multirow{2}*{Full Fine-tuned} & AOI & 89.78 & 94.27 & 91.98 & 81.25 & 78.93 & 95.54 & 99.94\\
  & AAI & 15.7 & 28.33 & 27.86 & 18.57 & 22.30 & 1.34 & 2.90\\
  \midrule
  \multirow{2}*{\emph{AT@stage-1}} & AOI & 86.02 & 92.31 & 86.23 & 61.87 & 70.48 & 95.78 & 99.94\\
  & AAI & 75.44 & 83.93 & 77.98 & 45.49 & 44.50 & 66.10 & 99.31\\
  \midrule
  \multirow{2}*{\emph{AT@stage-2}} & AOI & 40.28 & 91.55 & 90.55 & 70.38 & 70.35 & 80.68 & 99.92\\
  & AAI & 31.56 & 80.89 & 71.93 & 44.04 & 52.77 & 74.53 & 99.79\\
  \midrule
  \multirow{2}*{\emph{AT@stage-1\&2}} & AOI & 71.52 & 82.61 & 81.31 & 65.39 & 67.54 & 90.88 & 99.88\\
  & AAI & 64.49 & 76.68 & 77.28 & \textbf{59.02} & \textbf{58.48} & 86.93 & 99.83\\

  \midrule
  \multirow{2}*{\emph{KD@stage-2}} & AOI & 87.74 & 91.87 & 90.71 & 68.09 & 74.41 & 94.56 & 99.98\\
  & AAI & 21.37 & 31.97 & 42.06 & 4.75 & 5.43 & 44.28 & 89.50\\

  \midrule
  \multirow{2}*{\emph{MD@stage-1\&2}} & AOI & 86.48 & 91.71 & 87.17 & 64.83 & 70.04 & 95.62 & 99.96\\
  & AAI & \textbf{77.73} & \textbf{85.50} & \textbf{81.41} & 53.46 & 47.93 & \textbf{88.63} & \textbf{99.90}\\
 \bottomrule[1pt]
\end{tabular}}
\end{table*}

\subsection{Factors Influencing Robustness of Fine-tuned Model}

\paragraph{Model capacity.}
We then employ model size to examine the influence on fine-tuned model. Table~\ref{tab:model} shows the average results for $5$ ResNet-based backbones as pre-training architecture: ResNet-18 (RN-18), ResNet-50 (RN-50), ResNet-101 (RN-101), WideResNet-50-2 (WRN-50-2), and WideResNet-101-2 (WRN-101-2). It is easy to find that as network size increases, both the generalization and robustness consistently improve (with DR value decreasing from $81.47$ to $69.12$). This indicates that the high capacity of the pre-trained model alleviates the non-robustness transfer to the fine-tuned models.

\paragraph{Task difficulty.}

Task difficulty largely depends on the dataset. In this work, we measure task difficulty as the amount of semantics in the dataset necessary to solve the task. Specifically, we select $2$ source datasets for comparison: ImageNet-$10$animals (a subset of ImageNet, with sufficient semantics and containing images of animals) and CIFAR10 (with insufficient semantics and containing images of animals) with the equal number of training images ($50,000$ images of $10$ classes). To ensure the scale of source domain to target domain, we select $2$ target datasets that also contain images of animals: Pets and NICO. The performance of fine-tuning on different target datasets is reported in Table~\ref{tab:task}. It is unsurprising that employing ImageNet-$10$animals as pre-training dataset gives rise to fine-tuned models with higher AOI. However, the fine-tuned models transferred from CIFAR10 achieves lower DR (better robustness), which indicates that the source dataset with more semantics improves generalization yet has more risk to transfer non-robustness. Therefore, the guideline in selecting source dataset for robust fine-tuned models seems not that straightforward. Uncovering the paradox between generalization improvement and robustness decrease for pre-training needs to further study the mechanism of feature learning.

\section{Robust ImageNet Pre-training}

The related works have mainly focused on how to improve robustness of pre-training, but hardly any work has paid to how and why pre-training derives non-robustness. Ignoring it and simply using generic adversarial defenses may lead to a gap from the theoretically optimal robustness, while focusing on the difference between fine-tuned model and standard model has the potential to achieve better performance than the above generic adversarial defenses. Feature space steepness is a characterizing factor for robustness, and we observe that \emph{non-robust features are transferred resulting in steepening of the feature space}. In this section, we first introduce a metric to quantify the difference between target and source tasks, and then propose a method called \emph{Discrepancy Mitigating} that regularizes the steepness of the feature space at the two stages (denoted as \emph{DM@stage-1\&2}), and it outperforms most existing methods in transfer learning. So the significance of understanding why ImageNet pre-training transfer non-robustness goes beyond itself, and we consider DM@stage-1\&2 as \emph{a prompt (rather than the main contribution) for an open thread.}



\subsection{Steepness of Feature Space}
Since pre-training essentially serves as a feature extractor for the target task, we propose to measure the difference by examining how the features extracted from pre-trained model fit to the images of target task. Specifically, steepness of feature space is a recognized property closely related to model robustness~\cite{yang2020closer}. Local Lipschitzness (LL) is typically used to calculate steepness as following: 
\begin{equation}\label{eq1}
  {\rm LL}(f(X)) = \frac{1}{n} \sum_{i=1}^{n} \max_{x_{i}^{'} \in \mathbb{B}_{\infty}(x_i, \epsilon)}  \frac{\Vert f(x_i)-f(x'_i) \Vert_1}{\Vert x_i-x'_i \Vert_\infty},
\end{equation}
where $n$ denotes the number of images in dataset $X$, $x$ is original image from dataset $X$, and $x'$ is the corresponding adversarial image.

A lower value of LL implies smoother feature space which is usually with good robustness. We use ImageNet and Alphabet datasets as examples to respectively train pre-trained models $f^I(\cdot)$ and $f^A(\cdot)$, and then use them to extract features for images from Alphabet dataset $X^A$. ${\rm LLF}(f^I(X^A))$ and ${\rm LLF}(f^A(X^A))$ thus represent how ImageNet-trained and Alphabet-trained features fit to the Alphabet images. The result is ${\rm LL}(f^I(X^A))=367.4$ and ${\rm {\rm LL}}(f^A(X^A))=32.9$, indicating that the features pre-trained from ImageNet fail to fit to the Alphabet images. 

\subsection{Steepness Regularization}

We propose to reduce the steepness of pre-trained feature space on the target samples to mitigate the influence of the discrepancy between target and source tasks called Discrepancy Mitigating (inspired by a smooth representation-based defense~\cite{zhang2019theoretically}). Specifically, in addition to the traditional fine-tuning loss, LLF regularization term is added to derive the following objective function:
\begin{equation}\label{eq_method}
 \min_{\theta_f, \theta_g} \frac{1}{m} \sum_{i=1}^{m} \mathcal{C}(y, g(f(x_i))) + \lambda \cdot {{\rm LL}}(f(X)),
\end{equation}
where $\mathcal{C}$ is the cross-entropy classification loss, ${\rm LL}(f(X))$ is the steepness regularization term as defined in Equation~\eqref{eq1}, $x_i$ is original image from dataset $X$, $m$ denotes the number of images in dataset $X$, and $\lambda$ is the balancing parameter to control the trade-off between generalization and robustness. The hyperparameter $\lambda$ in this work is set to be $500$. The above optimization problem can be easily solved by PGD-like procedures.

To evaluate the effectiveness of steepness regularization in robust pre-training, we consider several baselines (listed in the Section~\ref{sec:related}) for comparison. Basically speaking, to improve the robustness of fine-tuned model involves with the two stages of fine-tuning and pre-training. Our proposed robust pre-training solution (denoted as \emph{DM@stage-1\&2}) combines the two stages: at the pre-training stage we employ adversarial training as in~\cite{DBLP:conf/nips/SalmanIEKM20} to obtain a robust pre-trained model, and at the fine-tuning stage we fine-tune on the target dataset according to Equation~\eqref{eq_method} to reduce the feature space steepness caused by the discrepancy between target and source tasks.


The experimental results of ResNet-18 backbone are shown in Table~\ref{tab:method_resnet18}. We have the following main findings: (1) Regarding robustness, \emph{MD@stage-1\&2} achieves superior AAI and DR in most examined datasets, owing to regularizing the transferred feature space steepness; (2) Regarding generalization, \emph{MD@stage-1\&2} guarantees performance compared to the original fine-tuned model, and achieves comparable if not better performance than the baseline methods. This demonstrates the feasibility of regularizing the difference between target and source tasks in addressing the paradox between robustness and generalization.

\section{Conclusion and Discussion}
\label{sec7}

\paragraph{Conclusion}
In this work, we demonstrate that ImageNet pre-training has risk to transfer non-robustness. We first found ImageNet pre-training not only transfer generalization into the fine-tuned model, but also the non-robustness. Then we discuss the reason for robustness decrease that the useful non-robust features for downstream tasks are transferred from pre-trained model. Therefore, from the perspective of the pre-trained model, we analyze the influencing factors of model capacity and task difficulty and further evaluate the impact on the fine-tuned model. Finally, we introduce a simple yet effective robust ImageNet pre-training solution.

\paragraph{Discussion}
This paper studies pre-training as the example paradigm of transfer learning. Also of vital importance is to examine the reliability of other transfer learning paradigms like knowledge distillation and domain adaption. A particular phenomenon is the non-reliability accumulation in iterative transfer learning. E.g., there has been an increasing attempt to automatically label data by a well-trained model~\cite{yalniz2019billion, zoph2020rethinking, xie2020self, kahn2020self}. Since it is difficult to tell whether the data is labeled by human or by model, there exists a risk to iteratively transfer the pseudo label from one to another model. Without human intervention to correct the potentially faulty knowledge, the continuous transfer of knowledge among models likely leads to the so-called ``echo chamber'' situation in sociology~\cite{barbera2015tweeting}. As observed in this work, one single round of knowledge transfer can contribute to considerable reliability issues, and iterative transfer may result in catastrophic results. In summary, many works remain to explore the mechanisms behind non-reliability transfer, and we are working towards developing more reliable transfer learning.

\section{Acknowledgments}
This work is supported by the National Natural Science Foundation of China (Grant No. 61832002, 62172094), and Beijing Natural Science Foundation (No.JQ20023).

\bibliography{aaai23}

\end{document}